\documentclass[runningheads]{llncs}
\usepackage[T1]{fontenc}
\usepackage{graphicx}
\usepackage{color}
\usepackage{booktabs}
\usepackage{cite}
\usepackage{hyperref}
\usepackage{amsmath}
\usepackage{amssymb}
\usepackage{svg}

\newcommand{\bftab}{\fontseries{b}\selectfont}

\begin{document}
\title{MAGDA: Multi-agent guideline-driven diagnostic assistance}
%
%
\author{David Bani-Harouni\inst{1,2} \and Nassir Navab\inst{1,2} \and Matthias Keicher\inst{1,2}}
\authorrunning{D. Bani-Harouni et al.}
%
\institute{Computer Aided Medical Procedures, School of Computation, Information and Technology, Technical University of Munich, Garching, Germany \and Munich Center for Machine Learning (MCML), Munich, Germany \\
\email{david.bani-harouni@tum.de}}
\maketitle              
\begin{abstract}
In emergency departments, rural hospitals, or clinics in less developed regions, clinicians often lack fast image analysis by trained radiologists, which can have a detrimental effect on patients’ healthcare. Large Language Models (LLMs) have the potential to alleviate some pressure from these clinicians by providing insights that can help them in their decision-making. While these LLMs achieve high test results on medical exams showcasing their great theoretical medical knowledge, they tend not to follow medical guidelines. In this work, we introduce a new approach for zero-shot guideline-driven decision support. We model a system of multiple LLM agents augmented with a contrastive vision-language model that collaborate to reach a patient diagnosis. After providing the agents with simple diagnostic guidelines, they will synthesize prompts and screen the image for findings following these guidelines. Finally, they provide understandable chain-of-thought reasoning for their diagnosis, which is then self-refined to consider inter-dependencies between diseases. As our method is zero-shot, it is adaptable to settings with rare diseases, where training data is limited, but expert-crafted disease descriptions are available. We evaluate our method on two chest X-ray datasets, CheXpert and ChestX-ray 14 Longtail, showcasing performance improvement over existing zero-shot methods and generalizability to rare diseases.

\end{abstract}
\section{Introduction}
Radiology holds a critical position in contemporary healthcare, being integral to the treatment and management of most patients. However, the healthcare sector is currently grappling with what has been termed the "radiologist shortage" \cite{rimmer2017radiologist}. In the UK, this shortage stands at 29\% and is predicted to worsen, reaching 40\% within the next four years \cite{radiologyreport}. This effect is exacerbated in rural hospitals or clinics in less developed regions of the world, where the population per radiologist is much greater \cite{ramli2023growing,vu2023growing}. When there is a lack of radiologists, the clinicians with patient contact have to either miss out on valuable radiological information or evaluate that information themselves without proper training. 
Large Language Models (LLMs) have recently demonstrated remarkable potential for reasoning and solving complex problems, presenting an opportunity to address this challenge~\cite{singhal2023large}. However, in a clinical context, deterministic models strictly adhering to evidence-based medical guidelines are preferred over creative but unpredictable LLM outputs. Moreover, generalist LLMs like GPT-4~\cite{achiam2023gpt} may lack domain-specific knowledge required for accurate diagnosis or have outdated medical insights. Consequently, providing LLMs access to relevant clinical knowledge sources, such as guidelines encapsulating the medical community's consensus, is critical for effective diagnostic processes, particularly for rare diseases with limited data.
In contrast to visual instruct tuning~\cite{NEURIPS2023_6dcf277e}, prompting Vision-Language Models (VLMs) is an intriguing approach to enabling LLMs to understand the content of images without the need to retrain. Recent explorations have successfully employed contrastive language-image pretraining (CLIP)~\cite{radford2021learning} for few- and zero-shot classification of common diseases in chest X-rays~\cite{seibold2022breaking,tiu2022expert,wang2022medclip,Windsor23,bannur2023learning,wu2023medklip}. Building on this, Xplainer~\cite{pellegrini2023xplainer} introduced a classification-by-description approach, querying a vision-language model for image observations indicative of a disease, providing inherent explainability. However, it naively averages concept probabilities, failing to account for dependencies between these concepts. To address those limitations, we propose MAGDA (Multi-Agent Guideline-driven Diagnostic Assistance), a multi-agent framework that unifies the incorporation of clinical guidelines as knowledge sources, dynamic prompting of a vision-language model for LLM understanding of radiology images, and a transparent diagnosis reasoning following the domain-specific knowledge provided by clinical guidelines. We show that this approach achieves state-of-the-art performance on zero-shot classification tasks of pathologies in the CheXpert dataset \cite{irvin2019chexpert} and rare diseases in the ChestXRay 14 Longtail dataset \cite{holste2022long}. Our key contributions are:

\begin{itemize}

\item An end-to-end guideline-driven approach that requires only a clinical guideline and a medical image as input to perform zero-shot diagnosis
\item Novel dynamic prompting of vision-language models to enable LLMs to screen medical images for unseen diseases without the need for fine-tuning
\item A transparent reasoning process through chain-of-thought reasoning, providing insights into the diagnostic decision-making

\end{itemize}

\section{Methodology}

\begin{figure}[t]
\includegraphics[width=\textwidth]{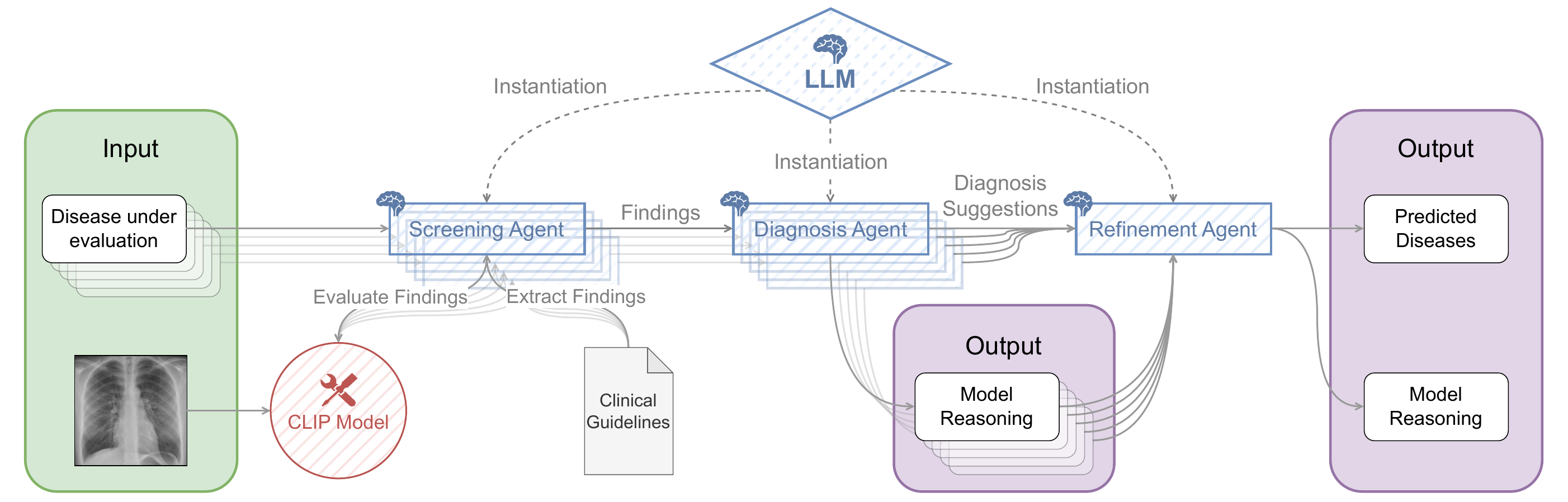}
\caption{Schematic overview of the proposed method MAGDA.} \label{fig:graph_abstract}
\end{figure}

\subsection{Model Overview}
We propose MAGDA, a multi-agent zero-shot method that can work with expert-crafted disease descriptions to provide transparent decision support. A general overview of the method is shown in Fig. \ref{fig:graph_abstract}. The LLMs used are not fine-tuned and all adaptions to the tasks are performed in-context. The multi-agent system consists of three agents that take over different tasks in the diagnosis procedure:
\begin{enumerate}
    \item \textbf{Screening agent $\mathcal{S}$:} This agent handles the image analysis. As an Augmented Language Model \cite{mialon2023augmented} with tool-using capabilities, it can prompt a CLIP model \cite{radford2021learning} to evaluate fine-grained image findings according to given diagnosis guidelines.
    \item \textbf{Diagnosis agent $\mathcal{D}$:} This agent is given the image findings from the screening agent and is tasked to reason about these findings to reach a diagnosis for the patient
    \item \textbf{Refinement agent $\mathcal{R}$:} This agent is responsible for refining the predictions of the diagnosis agent by considering inter-dependencies of diseases and evaluating the quality of the reasoning. It then gives the final diagnosis prediction for the patient at hand.
\end{enumerate}

\subsection{Screening Agent}
When diagnosing a specific patient $p \in P$, this agent is run once for every disease $d \in D$ that we want to evaluate. It is presented with expert-crafted fine-grained image findings and returns the positive or negative findings present in the image following the given diagnosis guidelines.
$$
\mathcal{S}^p_d(G_d, d) \rightarrow F_d^p,
$$
where $G_d$ are the disease guidelines, $d$ is the condition under evaluation, and $F_d^p$ are the patient and disease-specific positive and negative image findings.
The guidelines $G_d$ can be provided in either an already expert-crafted fine-grained list of disease-specific image findings or an unstructured disease description from which the model can extract these fine-grained image findings. For example, in the case of an enlarged cardiomediastinum, one image finding may be described as "Abnormal contour of the heart border".
In order to screen the image for the presence or absence of these image findings, we augmented the screening agent with the ability to prompt a CLIP model \cite{radford2021learning}. Following established works of classification-by-description \cite{tiu2022expert,seibold2022breaking,pellegrini2023xplainer}, we task the agent to use contrastive prompting, i.e., prompting the model with both a positive and a negative description. This has shown to be superior to just evaluating the similarity between the image embedding and the text embedding of the positive description. Since the findings are provided only in positive form, the model is tasked with creating negations following grammar rules and ensuring sensibility beyond simply appending the word "no" in front of the description. 
As we are not fine-tuning the LLM to be able to use the CLIP-tool, we provide an in-context description of the tool together with instructions on how to use it. Through one example, it is further tasked with following the report style template from Xplainer \cite{pellegrini2023xplainer}. During inference, the tool can be called using the call:
\begin{center}
    CLIP: [positive description] / [negative description] ->
\end{center}
 Once the descriptions have been extracted from the model output, we provide them to the CLIP model. Specifically, we employ the image and text encoder from BioVil-T \cite{bannur2023learning}. After computing the cosine similarity of the image embedding with the positive and negative text embeddings, we calculate the softmax over these two similarities to get the final probabilities of each finding. Initial results on the validation set showed that the BioVil-T model tends to over-predict positive findings, so we only count a positive finding if its probability exceeds a threshold $\psi$. Finally, depending on whether the tool returns a positive or negative result for the given description, we append "Positive" or "Negative" to the inference text and continue the LLM inference from there. After all given descriptions have been evaluated, the collected positive and negative disease descriptions are passed to the diagnosis agent.

\subsection{Diagnosis Agent}
The diagnosis agent is again run once per patient and condition under evaluation. It is given the list of findings extracted from the image by the screening agent and returns a positive or negative prediction, including the reasoning for that decision.
$$
\mathcal{D}^p_d(F_d^p, d) \rightarrow p_d^p, r_d^p,
$$
where $p_d^p$ is the binary disease prediction for patient $p$ and disease $d$, and $r_d^p$ is the reasoning for that prediction. 
It has been shown that LLM reasoning can be significantly improved by chain-of-thought prompting \cite{wei2022chain}, a prompt engineering technique where the model is asked to provide step-by-step reasoning before answering a question. Additionally, to an increase in reasoning capabilities, this reasoning makes the method inherently explainable. As the model provides explanations for its predictions, clinicians can use these explanations to evaluate the decision process and increase trust in the model output. Specifically, we ask the model to provide reasoning before answering the question "Does the patient have [$d$]?". We prompt the model to use a specified format to make parsing the model output possible, ending the reasoning process with the sentence: "Therefore, my answer is: [yes/no]."
Once the various predictions and reasonings have been collected, they are passed to the refinement agent for the final patient diagnosis.

\subsection{Refinement Agent}
The refinement agent is run once per patient. It is presented with all positive disease predictions and the diagnosis agent's reasonings for these predictions. It returns the final patient prediction.
$$
 \mathcal{R}^p(\{(p_d^p, r_d^p) | d \in D , p_d^p \text{ is positive} \}) \rightarrow \{\hat{p}_d^p | d \in D\} 
$$
The refinement agent is tasked with evaluating the provided reasoning. So far, every disease has been evaluated on its own in order to not overload the agents. At this step in the diagnosis process, inter-dependencies between diseases can be considered. The refinement agent is queried for every disease under evaluation if that disease is present or not and again asked to provide chain-of-thought reasoning for that decision. From the model replies we parse the final patient predictions $\hat{p}^p$. "No Finding" is predicted if all other disease predictions are negative.

\section{Experimental setup}
\subsubsection{Datasets and evaluation metrics}
We evaluate our method on two chest X-ray datasets, CheXpert \cite{irvin2019chexpert} and ChestXRay 14 Longtail \cite{wang2017chestx,holste2022long}. The CheXpert dataset includes manually annotated validation and test sets comprising 200 and 500 patients, respectively. It encompasses 14 different categories, featuring "No Finding", 12 pathology labels, e.g., "Pneumonia", and a class "Support Devices". On CheXpert, we perform multi-label classification. Most comparable methods evaluate using the Area Under the ROC-curve (AUC) metric. As our method generates discrete predictions, threshold-independent metrics, like AUC, cannot sensibly be evaluated. Instead, we report micro and macro F1-score, precision, and recall. The CLIP finding probability threshold $\psi$, which is used to combat the over-prediction of the CLIP model, is set to $0.55$ based on experiments on the validation set.

The ChestXRay14 Longtail dataset is an extension of the common ChestXRay 14 dataset by adding five additional disease findings, expanding the classification to 20 categories. These are divided into 7 head classes (most common), 10 medium classes (moderately common), and 3 tail classes (least common). The dataset includes a balanced validation and test set, each offering 15 or 30 images per class, respectively, to ensure comprehensive coverage and evaluation capabilities across the spectrum of conditions. We evaluate on this balanced test set with equal number of cases per class. Here, we perform single-label classification and report the accuracy on the three tail classes. In this setting, we prompt the CLIP model without description negation. As the screening and diagnosis agents always perform multi-label classification, we further adapt our refinement agent to decide on exactly one positive prediction.

\subsubsection{Implementation details}
The backbone of our method lies in a powerful LLM instantiated in different ways as the various agents. Unless stated otherwise, we employ the Mixtral 8x7B instruct model from Mistral AI \cite{jiang2024mixtral}. This model provides a good trade-off between memory efficiency, inference speed, and model capabilities. We use a 4-bit GPTQ quantized version of the model \cite{frantar2022gptq}. This reduces the memory requirements while keeping the loss in model accuracy minimal. Thus processed we are able to run text generations on a single Nvidia A40 GPU using a temperature of 0.8. Where available, we used the image findings from the public Xplainer repository\cite{pellegrini2023xplainer} as guidelines to ensure fair comparison. Where not available, we used a similar approach to Xplainer of prompting GPT-4 to generate candidate findings, and correcting them based on text book knowledge.

\section{Results and discussion}

\begin{table}[t]
    \centering
        \caption{Test set results for zero-shot classification on the CheXpert dataset. MAGDA (nG) is our proposed method without the use of guidelines, relying on LLM knowledge.}
    \begin{tabular}{lrrrrrr}
    \toprule
                &   \multicolumn{2}{c}{F1-score} & \multicolumn{2}{c}{Precision} & \multicolumn{2}{c}{Recall} \\
    \cmidrule(lr){2-3}
    \cmidrule(lr){4-5}
    \cmidrule(lr){6-7}
       Method  & micro & macro & micro & macro & micro & macro \\
       \midrule
       CheXzero & 35.69 & 33.50 & 25.58 & \bftab37.72 & 58.98 & 64.88 \\  
       Xplainer & 45.33 & 39.27 & 31.36 & 33.74 & 81.74 & 83.27 \\
       MAGDA (nG) & 42.94 & 36.50 & 29.39 & 30.82 & 79.62 & 82.07 \\
       \bftab{MAGDA} & \bftab46.18 & \bftab39.58 & \bftab31.93 & 33.43 & \bftab83.43 & \bftab83.47 \\
       \bottomrule
    \end{tabular}
    \label{tab:test_results_CheXpert}
\end{table}

\begin{table}[t]
    \centering
    \caption{Test set results on the ChestXray 14 Longtail dataset. Accuracy is the classification accuracy on the rare tail classes. Methods above the line are fully supervised.}
    \begin{tabular}{llr}
    \toprule
       Method  & Zero-shot & Accuracy \\
       \midrule
       ResNet-50 \cite{holste2022long} & $\times$ & 1.7 \\
       Decoupling-cRT \cite{holste2022long} & $\times$ &  30.0 \\
       \midrule
       CheXzero & $\checkmark$ &  12.3  \\  
       Xplainer & $\checkmark$ &  8.2  \\
       \bftab{MAGDA} & $\checkmark$ & \bftab18.5  \\
       \bottomrule
    \end{tabular}
    
    \label{tab:test_results_CXR14_LT}
\end{table}

\begin{table}[t]
    \centering
        \caption{Comparison of zero-shot classification of the diagnosis agent, i.e., before refinement with the refinement agent, using different LLMs as a reasoning backbone on the test set of the CheXpert dataset.}
    \begin{tabular}{lrrrrrr}
    \toprule
                &   \multicolumn{2}{c}{F1-score} & \multicolumn{2}{c}{Precision} & \multicolumn{2}{c}{Recall} \\
    \cmidrule(lr){2-3}
    \cmidrule(lr){4-5}
    \cmidrule(lr){6-7}
       Reasoning Model  & micro & macro & micro & macro & micro & macro \\
       \midrule
       GPT-4 & \bftab46.87 & \bftab41.10 & \bftab32.17 & \bftab34.17 & 86.31 & 85.13 \\  
       Llama2 70B chat & 46.36 & 40.72 & 31.62 & 33.48 & \bftab86.87 & \bftab88.01 \\
       Mixtral 8x7B instruct & 45.31 & 39.70 & 30.85 & 33.41 & 85.25 & 86.62 \\
       \bottomrule
    \end{tabular}

    \label{tab:test_CheXpert_reason_models}
\end{table}

In Table \ref{tab:test_results_CheXpert}, we compare with state-of-the-art zero-shot classification methods CheXzero \cite{tiu2022expert} and Xplainer \cite{pellegrini2023xplainer} on the CheXpert test dataset. Because CheXzero is only evaluated on the six competition pathologies in the original paper, we use their public code and model to evaluate on all CheXpert classes. Most state-of-the-art methods only report the AUC, we can therefore only compare with methods with published code and calculate their respective F1-score, precision, and recall. For example, a comparison with Seibold et al. \cite{seibold2022breaking} or ELIXR \cite{xu2023elixr} was not possible for that reason. We outperform all comparable methods on zero-shot classification on all metrics except macro precision, where CheXzero has a better score at the expense of a much lower recall. Here, we also compare the guideline-driven approach with the generation of findings by the model itself, showing that the provision of guidelines to our method increases performance. 
Table \ref{tab:test_results_CXR14_LT} shows a comparison on the ChestXray14 Longtail dataset with the same zero-shot methods as before and additionally with two fully supervised methods trained on that dataset \cite{holste2022long}. Here, we again outperform both zero-shot methods on the tail classes. Notably, we even reach a higher accuracy than a simple supervised method trained on the 68,058 training samples. Only a highly tuned method employing decoupled training \cite{holste2022long} achieves a higher accuracy. These results show that the provision of detailed guidelines describing the diagnosis of rare and lesser-known diseases can help with diagnostic accuracy.
In Table \ref{tab:test_CheXpert_reason_models}, we compare exchanging our Mixtral 8x7B instruct model for other well-known LLMs, namely Llama 2 70B chat \cite{touvron2023llama} and GPT-4 \cite{achiam2023gpt}. While both alternative models reach higher performance, this comes at the cost of higher computational needs in the case of Llama2 70B chat, or dependence on a proprietary API. 

\subsubsection{Ablation studies}

We now want to look at the benefits of different aspects of our method. First, in Table \ref{tab:negation_ablation}, we compare the rule-based negation by simply appending a "no" before the finding description with the LLM-created finding negation done by the screening agent. We see that the latter results in better performance, highlighting the benefit of descriptions that are more aligned with natural language and thus the style of radiology reports. 
In Table \ref{tab:refinement_ablation}, different approaches for the refinement agent are compared. We compare combinations of chain-of-thought reasoning and including the CheXpert disease graph in textual form \cite{irvin2019chexpert}. This disease graph models the dependencies between classes, e.g. "Enlarged Cardiomediastinum" being a sign of "Cardiomegaly". However, both in the case of using chain-of-thought reasoning and not using it, the inclusion of the disease graph decreases the performance.

\begin{table}[t]
\centering
\begin{minipage}[]{.45\textwidth}
\centering
\caption{Comparison of naive finding negation with the improved negation by the Screening agent. Results are reported before the refinement by the refinement agent on the CheXpert validation set.}
\begin{tabular}{lrr}
\toprule
& \multicolumn{2}{c}{F1-score} \\
\cmidrule(lr){2-3}
CLIP prompting &  micro & macro \\
\midrule
Naive negation &  46.79 & 41.75 \\
LLM negation &  \bftab48.00 & \bftab42.10 \\
\bottomrule
\end{tabular}

\label{tab:negation_ablation}
\end{minipage}\hspace*{0.7cm}
\begin{minipage}[]{.45\textwidth}
\centering
\caption{Comparison of different refinement approaches on the CheXpert validation set. CoT = chain-of-thought reasoning, DG = disease graph.}
\begin{tabular}{llrr}
\toprule
& & \multicolumn{2}{c}{F1-score} \\
\cmidrule(lr){3-4}
CoT & DG & micro & macro \\
\midrule
$\times$ & $\times$ & 48.10 & 41.56 \\
$\times$ & $\checkmark$ & 47.79 & 40.39 \\
$\checkmark$ & $\times$ & \bftab49.17 & \bftab42.05 \\
$\checkmark$ & $\checkmark$ & 47.52 & 40.70 \\
\bottomrule
\end{tabular}

\label{tab:refinement_ablation}
\end{minipage}

\end{table}

\subsubsection{Qualitative results}
\begin{figure}[t]
\includegraphics[trim=0 0.5cm 0 0.5cm, clip, width=\textwidth]{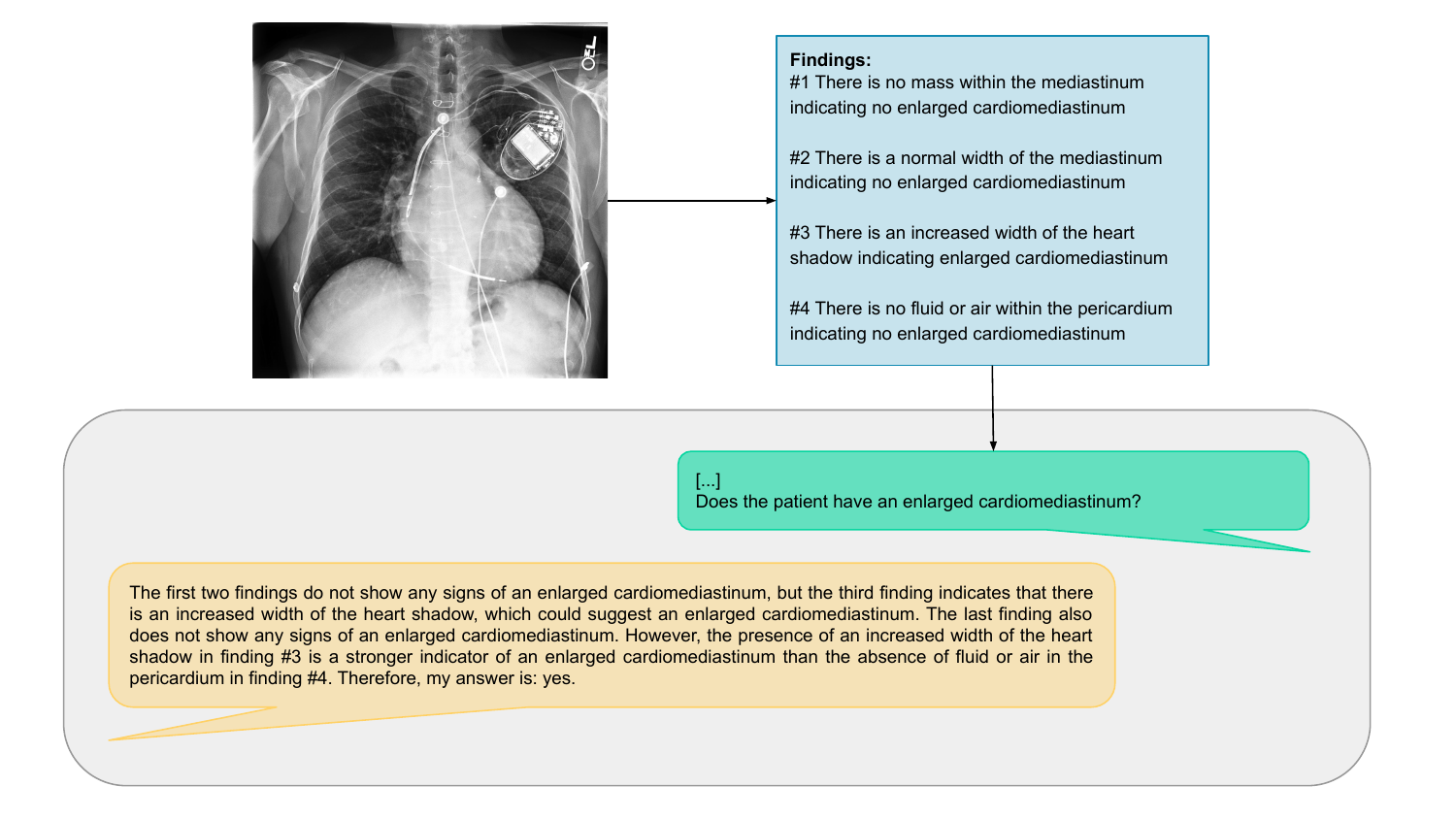}
\caption{A qualitative example of the model reasoning.} \label{fig:qualitative}
\end{figure}

In Fig. \ref{fig:qualitative}, we show a qualitative example of the reasoning provided by the diagnosis agent. The agent is presented with conflicting findings regarding the diagnosis of an enlarged cardiomediastinum. Instead of naively aggregating them, giving each finding the same importance, the diagnosis agent can differentiate between more and less relevant findings and can employ reason to reach a diagnosis. This example also shows the high dependence on the correctness of the image findings identified by the CLIP model. Even perfect reasoning cannot reach the correct conclusion if it works with wrong information. Therefore, improvements in the performance of VLMs can directly translate into further improvement of our method.

\section{Conclusion}
In this paper, we presented MAGDA, a novel multi-agent approach that integrates clinical guidelines, dynamic vision-language model prompting, and large language model reasoning to address the challenges of diagnostic assistance using LLMs. Our approach leverages the strengths of both LLMs and VLMs, enabling the zero-shot classification of diseases without the need for model retraining or fine-tuning. This guideline-driven methodology not only facilitates accurate diagnoses from medical images but also introduces a transparent reasoning process that enhances the explainability and trustworthiness of the diagnostic outcomes. Our evaluation on the CheXpert and ChestXray14 LT datasets demonstrates the effectiveness of our approach, particularly in scenarios involving rare diseases where traditional diagnostic methods are often hindered by data scarcity. By incorporating domain-specific knowledge through clinical guidelines and employing dynamic prompting techniques, we improve diagnostic accuracy and model trustworthiness.

\begin{credits}
\subsubsection{\ackname} The authors gratefully acknowledge the financial support by the Bavarian Ministry of Economic Affairs, Regional Development and Energy (StMWi) under project ThoraXAI (DIK-2302-0002).
\end{credits}
%
%
%
\bibliographystyle{splncs04}
\bibliography{bibliography}

\end{document}


\section*{MAGDA Supplementary Material}
\section{Agent prompts}
\subsection{Screening agent}
You are a screening agent who has to evaluate a chest x-ray image for the presence of certain image findings indicative of the presence or absence of <condition>. You can check for these findings using a tool called CLIP-Image-Analyzer (CLIP). When you query this tool with a description of an image finding and it's negation, the tool will return if that description matches the image.
Here is an example of how you can use the tool:

CLIP: There is an increased width of the heart shadow indicating an enlarged cardiomediastinum. / There is a normal heart shadow indicating no enlarged cardiomediastinum. -> Positive

Before the slash you provide the positive finding description indicating the disease and after the slash the negation of that finding indicating the absence of the disease. Do not forget to end your CLIP call with the arrow symbol "->". The tool will return if the positive finding matches the image or the negative one.
Here are the image findings you have to test for:
<xplainer\_findings> 

Follow the exact format given above. You are only given the positive finding, so you have to construct the negation yourself. Start the finding descriptions with "There is" or "There are" and end them with "indicating <condition>" or "indicating no <condition>". Evaluate all the exact image findings given above. Do not skip any and do not add any. Do not give a diagnosis yourself.

\subsection{Diagnosis Agent}
You are a diagnosis agent, that decides whether a patient has a specific condition given a list of the presence or absence of different image findings. Even if the diagnosis is not clear, you have to provide your best guess to answer the question. Provide your reasoning for your decision first, then answer the question. Be short and concise. End your reasoning with one of these exact sentences: \"Therefore, my answer is: yes." or "Therefore, my answer is: no.".

Here are the findings:
<findings>

Question: Does the patient have <condition>?

\subsection{Refinement Agent}
You are a medical expert who has to evaluate a number of diagnoses for a patient regarding consistency and reasonableness. The patient may have none, one or multiple of the following conditions:
<condition\_list>

You are given a list of positive diagnoses and reasonings for these diagnoses. You have to evaluate whether the reasoning is correct and the various diagnoses are consistent with each other. You are then asked for each condition individually if you think the patient has this condition. Each time first provide your short and concise reasoning, then end your reasoning with one of these exact sentences: "Therefore, my answer is: yes." or "Therefore, my answer is: no."

Here are the positive diagnoses for this patient:
<diagnoses>

If you understand your task, reply with “OK.” only.